\documentclass[journal]{IEEEtran}
\usepackage{times}
\usepackage{graphicx}
\usepackage{amsmath}
\usepackage{amssymb}
\usepackage{multirow}
\usepackage{breqn}
\usepackage{algorithm}
\usepackage{algorithmic}
\usepackage{array}
\ifCLASSOPTIONcompsoc
\usepackage[nocompress]{cite}
\else
\usepackage{cite}
\fi

\usepackage{booktabs}
\usepackage{subcaption}
\usepackage{graphbox}
\usepackage{arydshln}
\usepackage{authblk}
\usepackage[numbers]{natbib} 
\usepackage{helvet} 
\usepackage{courier}  
\usepackage{ wasysym }

\newcommand{\figintroImg}{
\begin{figure}[t]
  \centering
  \includegraphics[width=\linewidth]{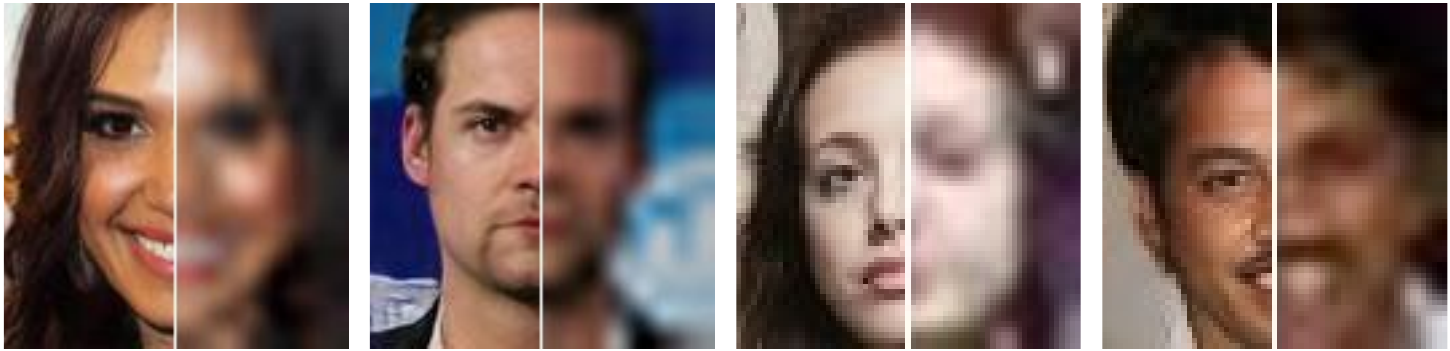}
  \caption{
  Visual samples of our face super-resolution method.
  Two images in the left are results on the CelebA dataset with 8x upsampling scale, while the right two are from real-world images with 4x upsampling scale.
  }
  \label{fig:introImg}
\end{figure}
}

\newcommand{\figpipeline}{
\begin{figure}[t]
  \centering
  \begin{tabular}{c}
    \begin{minipage}{0.95\linewidth}
    \centering
    \includegraphics[width=\linewidth]{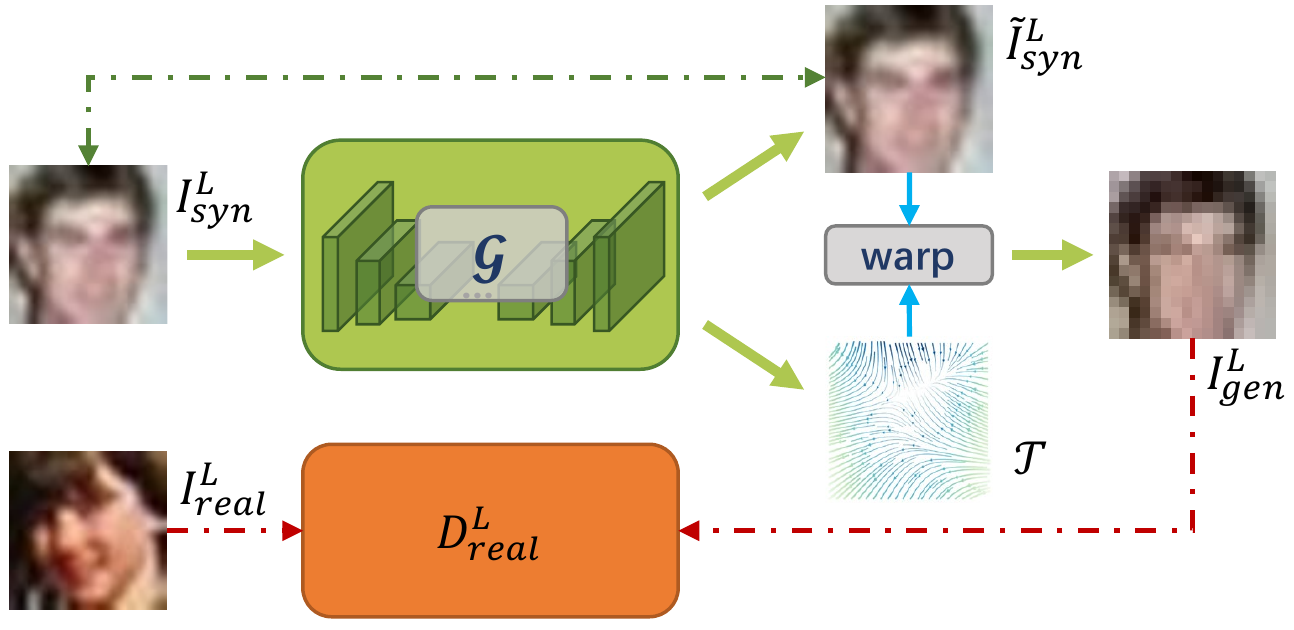}
    \subcaption{Domain adaptation in low-resolution space.}
    \label{fig:two_approaches_a}
    \end{minipage}\vspace{2mm}\\

    \begin{minipage}{0.95\linewidth}
    \centering
    \includegraphics[width=\linewidth]{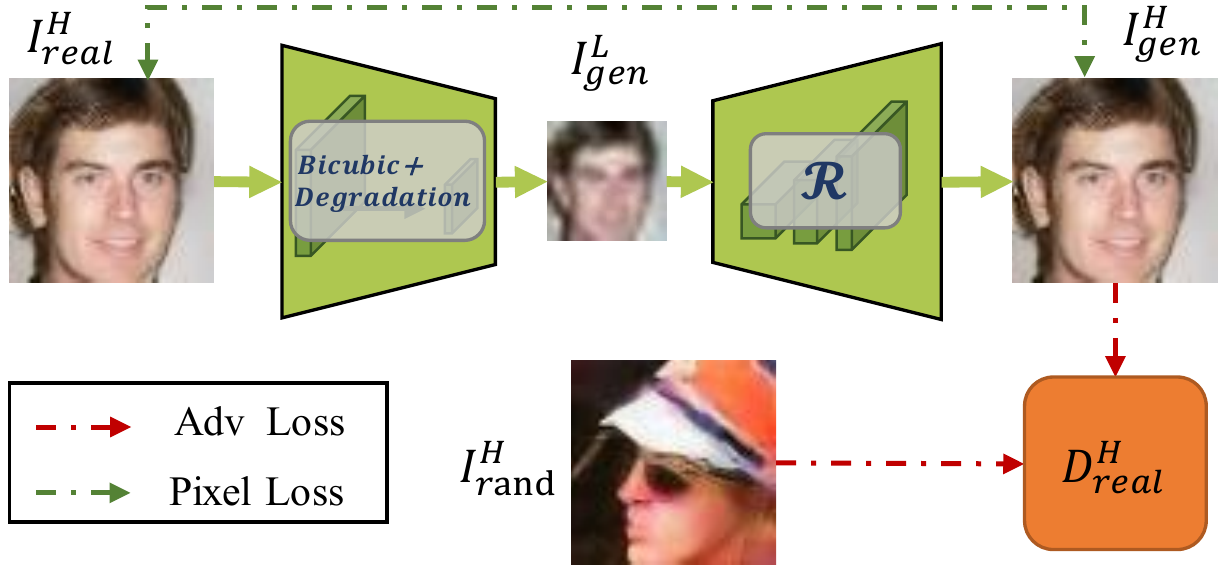}
    \subcaption{Super-resolution with pseudo pairs.}
    \label{fig:two_approaches_b}
    \end{minipage}
  \end{tabular}
  \caption{The pipeline of our two-stage unsupervised face super-resolution framework. 
  (a) The domain adaptation stage is to transfer the synthesized LR images (by bicubic downsampling) to its noisy counterparts which follow the real LR distribution. 
  (b) Training super-resolution network $\cR(\cdot)$ with pseudo pairs $(I_{gen}^{L}, I_{real}^{H})$ to adapt real-world degradation.}
  \label{fig:pipeline}
\end{figure}
}

\newcommand{\figgenerator}{
\begin{figure}[t]
  \centering
  \includegraphics[width=\linewidth]{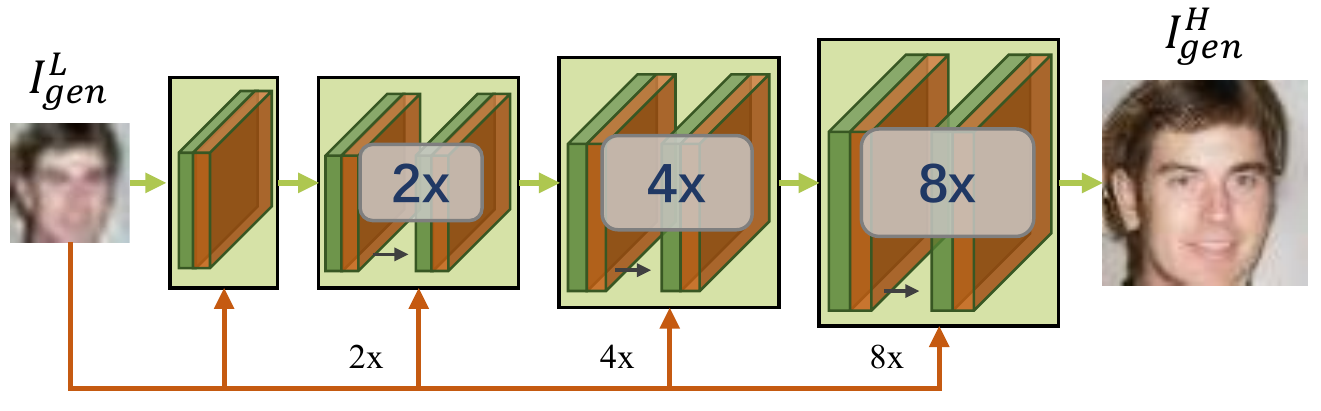}
  \caption{Self-conditioned SR network takes a degraded LR image ${I}_{gen}^{L}$ as input and gradually upscales it to the HR counterpart, along with corresponding bicubic upsampled image as condition.
  Green and orange cubes indicate convolution layer and instance normalization layer, respectively.}
  \label{fig:generator}
\end{figure}
}

\newcommand{\figSelfConditionedBlock}{
\begin{figure}[t]
  \centering
  \includegraphics[width= 0.9\linewidth]{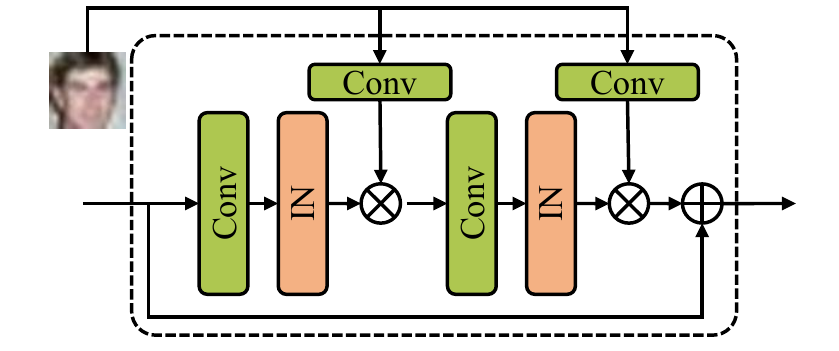}
  \caption{Self-conditioned block takes the input image as condition term, which is utilized to denormalize deep features. The activation layer is omitted.}
  \label{fig:self_conditioned_block}
\end{figure}
}

\newcommand{\figcomparisonSR}{
\begin{figure*}[t]
  \addtolength{\tabcolsep}{-5pt}
  \centering
  \resizebox{\linewidth}{!}{\begin{tabular}{cccccccc}
    \includegraphics[width=0.123\textwidth]{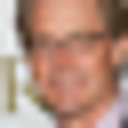}&
    \includegraphics[width=0.123\textwidth]{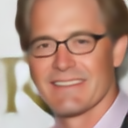}&
    \includegraphics[width=0.123\textwidth]{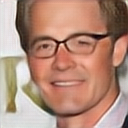}&
    \includegraphics[width=0.123\textwidth]{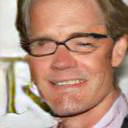}&
    \includegraphics[width=0.123\textwidth]{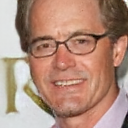}&
    \includegraphics[width=0.123\textwidth]{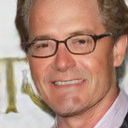}&
    \includegraphics[width=0.123\textwidth]{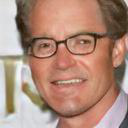}&
    \includegraphics[width=0.123\textwidth]{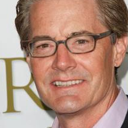}\\
    
    \includegraphics[width=0.123\textwidth]{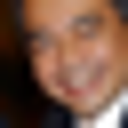}&
    \includegraphics[width=0.123\textwidth]{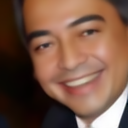}&
    \includegraphics[width=0.123\textwidth]{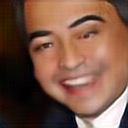}&
    \includegraphics[width=0.123\textwidth]{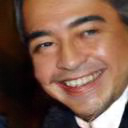}&
    \includegraphics[width=0.123\textwidth]{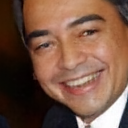}&
    \includegraphics[width=0.123\textwidth]{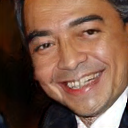}&
    \includegraphics[width=0.123\textwidth]{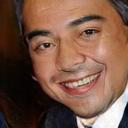}&
    \includegraphics[width=0.123\textwidth]{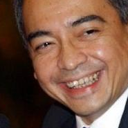}\\
    
    \includegraphics[width=0.123\textwidth]{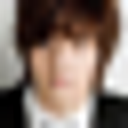}&
    \includegraphics[width=0.123\textwidth]{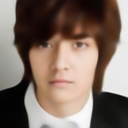}&
    \includegraphics[width=0.123\textwidth]{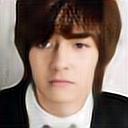}&
    \includegraphics[width=0.123\textwidth]{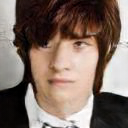}&
    \includegraphics[width=0.123\textwidth]{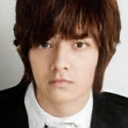}&
    \includegraphics[width=0.123\textwidth]{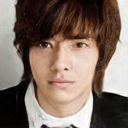}&
    \includegraphics[width=0.123\textwidth]{figures/compare/compare_CelebA_2_Our_pair.png}&
    \includegraphics[width=0.123\textwidth]{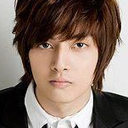}\\
    
    \includegraphics[width=0.123\textwidth]{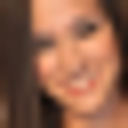}&
    \includegraphics[width=0.123\textwidth]{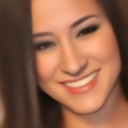}&
    \includegraphics[width=0.123\textwidth]{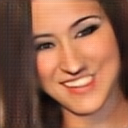}&
    \includegraphics[width=0.123\textwidth]{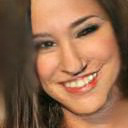}&
    \includegraphics[width=0.123\textwidth]{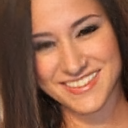}&
    \includegraphics[width=0.123\textwidth]{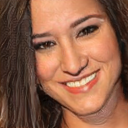}&
    \includegraphics[width=0.123\textwidth]{figures/compare/compare_CelebA_3_Our_pair.png}&
    \includegraphics[width=0.123\textwidth]{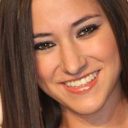}\\
    
    \includegraphics[width=0.123\textwidth]{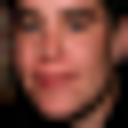}&
    \includegraphics[width=0.123\textwidth]{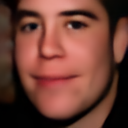}&
    \includegraphics[width=0.123\textwidth]{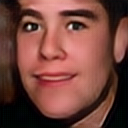}&
    \includegraphics[width=0.123\textwidth]{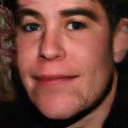}&
    \includegraphics[width=0.123\textwidth]{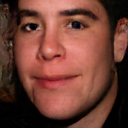}&
    \includegraphics[width=0.123\textwidth]{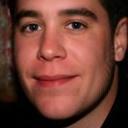}&
    \includegraphics[width=0.123\textwidth]{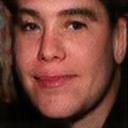}&
    \includegraphics[width=0.123\textwidth]{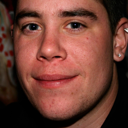}\\
    
    \includegraphics[width=0.123\textwidth]{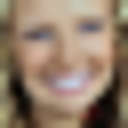}&
    \includegraphics[width=0.123\textwidth]{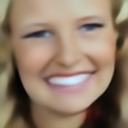}&
    \includegraphics[width=0.123\textwidth]{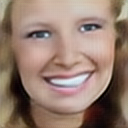}&
    \includegraphics[width=0.123\textwidth]{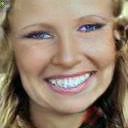}&
    \includegraphics[width=0.123\textwidth]{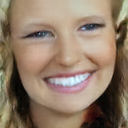}&
    \includegraphics[width=0.123\textwidth]{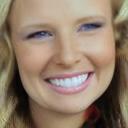}&
    \includegraphics[width=0.123\textwidth]{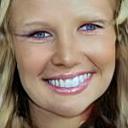}&
    \includegraphics[width=0.123\textwidth]{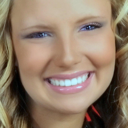}\\
    
    Bicubic & DIC~\cite{ma2020deep} & FSRGAN~\cite{chen2018fsrnet} & PFSR~\cite{kim2019progressive} 
    & DICGAN~\cite{ma2020deep} & Ours-paired & Ours-unpaired & HR \\
  \end{tabular}}
  \caption{Qualitative comparisons with state-of-the-art FSR methods, where each row shows the results for one randomly selected person with different methods. 
  The first column shows results with vanilla bicubic, second to fifth columns are from different SOTA methods, columns six through seven show our results with/without paired images, and the last column is the high-resolution ground truth. 
  Best viewed on screen for more details.}
  \label{fig:comparisonSR}
\end{figure*}
}

\newcommand{\figcomparisonRealSR}{
\begin{figure}[t]
  \centering
  \includegraphics[width=\linewidth]{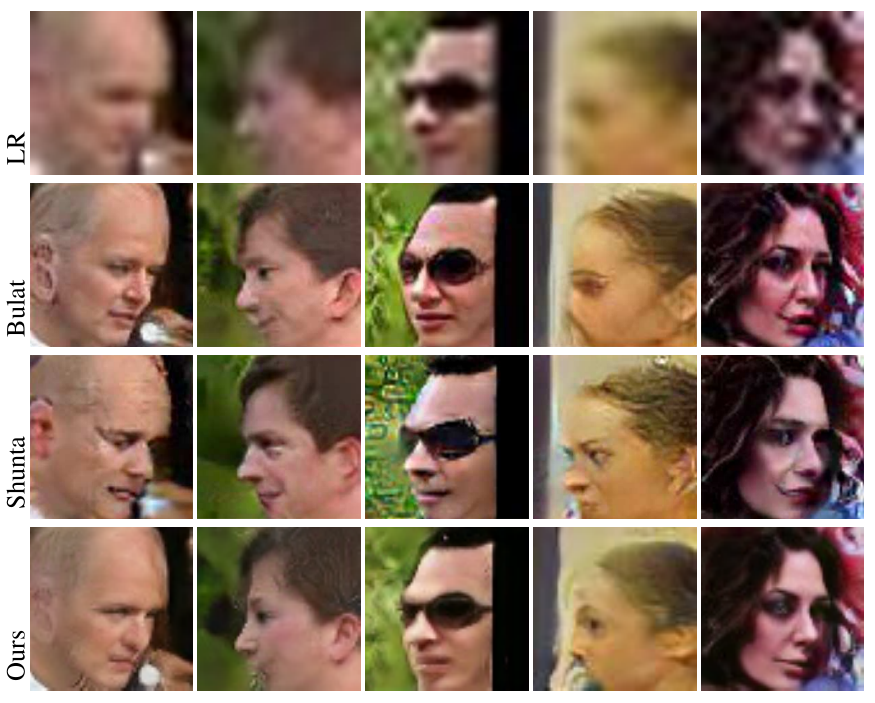}
  \caption{Qualitative comparisons with state-of-the-art FSR methods on real-world face super-resolution. 
  Each column shows the results for one randomly selected person in real world with different methods.}
  \label{fig:comparisonRealSR}
\end{figure}
}

\newcommand{\figablation}{
\begin{figure}[t]
  \centering
  \includegraphics[width=\linewidth]{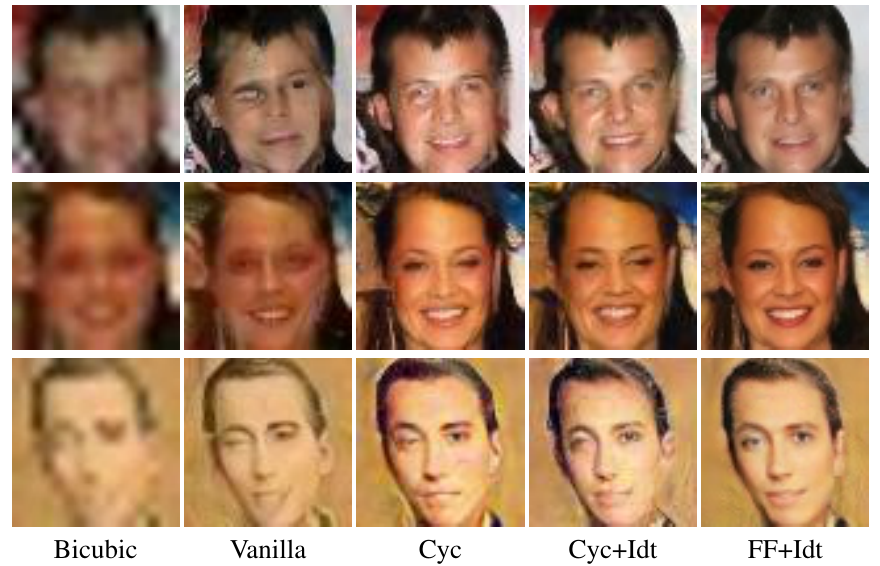}
  \caption{SR results of different variations on real LR faces. The abbreviation is same as Table~\ref{tab:ablation}.}
  \label{fig:comparisonAblation}
\end{figure}
}

\newcommand{\figDisturb}{
\begin{figure}[t]
  \centering
  \includegraphics[width=\linewidth]{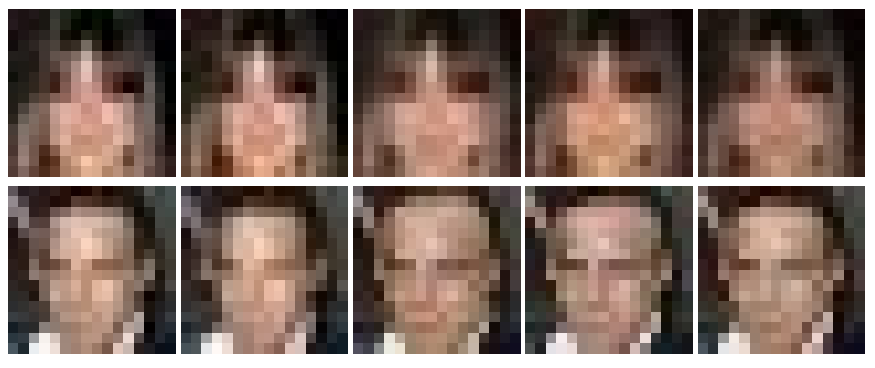}
  \caption{Visual results of flow field disturbance. The first column is downsampled LR images, the second is noisy LR images estimated by degradation network, and the others are generated by randomly disturbing estimated flow field.}
  \label{fig:disturb}
\end{figure}
}

\newcommand{\figfailure}{
\begin{figure}[t]
  \centering
  \includegraphics[width=\linewidth]{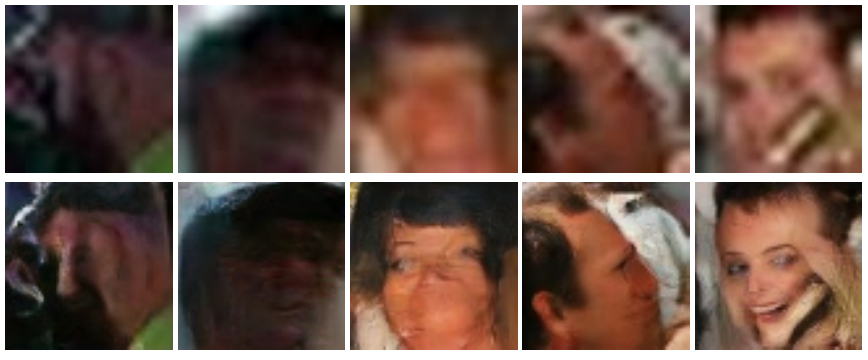}
  \caption{Some visual samples of typical failure case generated by our method.
  The first row represents different typical low-resolution samples, while the second row shows corresponding generated high-resolution images.}
  \label{fig:failure}
\end{figure}
}

\newcommand{\figVisual}{
\begin{figure}[t]
  \centering
  \includegraphics[width=\linewidth]{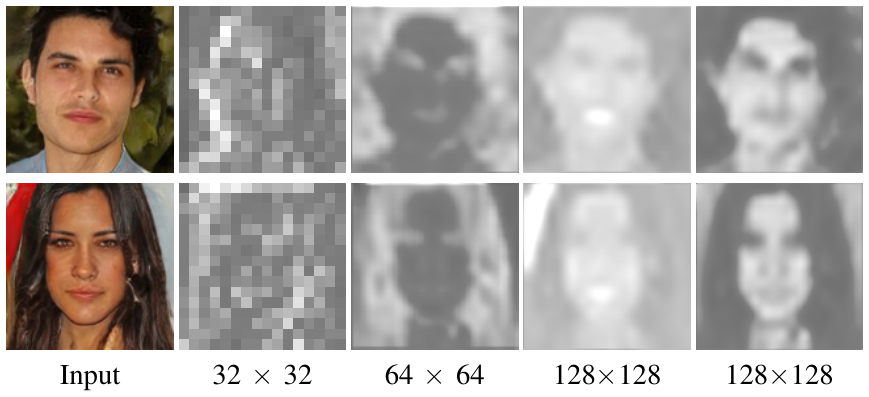}
  \caption{
    Visualization of some feature maps in condition branch of the self-conditioned block.
    The brighter the area, the larger the activation value.
    }
  \label{fig:visual}
\end{figure}
}

\newcommand{\ud}[1]{\underline{#1}}

\newcommand{\tabcomparisonSR}{
\begin{table*}[t]
  \centering
  \caption{Quantitative comparisons with SOTA methods on CelebA and Helen datasets in several commonly used metrics. Ours-P adopts paired images during the training phase while Ours-UP uses unpaired images. 
  L represents that extra facial landmark is involved in the training stage. 
  The bold and underscore indicate optimal and sub-optimal results, respectively. 
  The upward arrow indicates that the higher the value, the better, and vice versa.}
  \label{tab:comparisonSR}
  \resizebox{\linewidth}{!}{
  \addtolength{\tabcolsep}{1pt}
  \begin{tabular}{lc cccc cccc c}
    \toprule
    \multirow{3}{*}{Method} & \multirow{3}{*}{L}& \multicolumn{4}{c}{CelebA} & \multicolumn{4}{c}{Helen} & \multirow{3}{*}{\#Param.} \\
    \cmidrule(lr){3-6} \cmidrule(l){7-10}
    & &   PSNR $\uparrow$ &   SSIM $\uparrow$ &   LPIPS $\downarrow$  &   FID $\downarrow$ &   
          PSNR $\uparrow$ &   SSIM $\uparrow$ &   LPIPS $\downarrow$  &   FID $\downarrow$ \\
    \midrule
    Bicubic & 
    &   23.58 &   0.6285 &   0.5219 &   159.06 
    &   23.89 &   0.6751 &   0.5026 &   171.79 &   - \\
    RDN~\cite{zhang2018residual} & 
    &   26.13 &   0.7412 &   0.3135 &   96.14  
    &   25.34 &   0.7249 &   0.3496 &   104.38 &   24.34 M\\
    FSRNet~\cite{chen2018fsrnet} & \checkmark 
    &   26.48 &   \ud{0.7718} &   0.2968 &   94.39 
    &   25.90&   0.7759&   0.3539 &   101.13 &   2.15 M\\
    DIC~\cite{ma2020deep}     & \checkmark 
    &  \textbf{27.37} &   \textbf{0.7962} &   0.2565 &   89.24 
    &   \ud{26.69} &   \ud{0.7933} &   0.3378 &   94.34 &   21.80 M\\
    FSRGAN~\cite{chen2018fsrnet} & \checkmark 
    &  25.06 &   0.7311 &    0.2342 &   85.17 
    &   24.99 &  0.7424 &   0.2799 &   82.21 &   2.15 M\\
    PFSR~\cite{kim2019progressive} & \checkmark 
    &  24.43 &   0.6991 &   0.1500 &   69.33 
    &   24.73 &   0.7323 &   0.2017 &   77.62 &   8.97 M \\
    DICGAN~\cite{ma2020deep} & \checkmark 
    &  26.34 &    {0.7562} &   \ud{0.0840} &    \ud{34.49} 
    &    {25.96} &    0.7624 &   \ud{0.0917} &    \ud{50.21} &    21.80 M\\
    \midrule
    Ours-P &  
    &    \ud{26.59} &   0.7696 &    \textbf{0.0726} &   \textbf{30.25} 
    &   \textbf{26.87} &   \textbf{0.7962} &    \textbf{0.0911} &   \textbf{50.03} &   17.89 M \\
    Ours-UP &  
    &   23.42 &   0.6539 &   0.1141 &   35.61 
    &   23.59 &   0.6691 &   0.1346 &   67.53 &   17.89 M\\
    \bottomrule
  \end{tabular}}
\end{table*}}

\newcommand{\tabcomparisonRealSR}{
\begin{table}[t]
  \centering
  \setlength\tabcolsep{25pt}
  \fontsize{9}{12}\selectfont
  \caption{Quantitative results on real-world face super-resolution. 
  }
  \label{tab:face_sota}

  \begin{tabular*}{0.6\columnwidth}{@{}lr@{}}
    \toprule
    Method                                &  FID $\downarrow$ \\
    \midrule
    DeepDeblur~\cite{nah2017deep}         &  294.96        \\
    {Wavelet-SRNet~\cite{huang2017wavelet}} &  149.46   \\
    {SRGAN~\cite{ledig2017photo}}         &  126.75  \\
    {FSRNet~\cite{chen2018fsrnet}}        &  148.97   \\
    {CycleGAN~\cite{zhu2017unpaired}}     &  37.63  \\
    {Bulat~\etal~\cite{bulat2018learn}}   &  26.41     \\
    {Shunta~\etal~\cite{maeda2020unpaired}}  &  33.19   \\
    \midrule
    {Our} &   \textbf{25.07} \\
    \bottomrule
  \end{tabular*}
\end{table}
}

\newcommand{\tabablationSR}{
\begin{table}[ht]
  \centering
  \setlength\tabcolsep{5pt}
  \fontsize{9}{12}\selectfont
  \caption{Ablations for components in self-conditioned SR network on CelebA. 
  Top is the results under the supervised setting while the bottom is unsupervised. 
  \textbf{SCB}: Self-conditioned block.
  \textbf{Prog.}: Progressive training.
  \textbf{R1}: R1 regularization.
  }
  \label{tab:ablation_SR}

  \begin{tabular*}{1\columnwidth}{@{}cccc cccc@{}}
    \toprule
      Vanilla &   SCB &   Prog. &   R1 &   PSNR $\uparrow$ &   SSIM $\uparrow$ &   LPIPS $\downarrow$  &   FID $\downarrow$ \\
    \midrule
      \checkmark &  &   \checkmark &   \checkmark &   25.57 &   0.7354 &   0.1105  &   37.12 \\
     &   \checkmark  &  &   \checkmark  &   25.83 &   0.7426 &   0.1023  &   34.97 \\
      &   \checkmark  &   \checkmark &  &   25.94 &   0.7483 &   0.0912  &   33.46 \\
     &   \checkmark &   \checkmark &   \checkmark &   26.59 &   0.7696 &   0.0726  &   30.25 \\
    \midrule
      \checkmark &  &   \checkmark &   \checkmark &   22.72 &   0.6101 &   0.1523  &   56.18 \\
     &   \checkmark  &  &   \checkmark  &   22.88 &   0.6135 &   0.1428  &   52.67 \\
      &   \checkmark  &   \checkmark &  &   23.64 &   0.6660 &   0.1238  &   39.09 \\
     &   \checkmark &   \checkmark &   \checkmark &   23.42 &   0.6539 &   0.1141  &   35.61 \\
    \bottomrule
  \end{tabular*}
\end{table}
}

\newcommand{\tabablation}{
\begin{table}[t]
  \centering
  \setlength\tabcolsep{15pt}
  \fontsize{9}{12}\selectfont
  \caption{Ablation study for domain adaptation on real-world degradation. 
  \textbf{Vanilla}: train SR network with clean LR-HR pairs and test it on real LR images. 
  \textbf{Cyc}: perform domain adaptation with CycleGAN~\cite{zhu2017unpaired}. 
  \textbf{FF}:  flow field degradation network. 
  \textbf{Idt}: identity loss in Eq.(~\ref{eq:idt_loss}).
  FID is calculated between the final SR results and real HR images.
  FID$^\ddagger$ denotes the distance between the estimated LR images and the real ones.}
  \label{tab:ablation}

  \begin{tabular*}{1\columnwidth}{@{}ccccrr@{}}
    \toprule
    Vanilla  & Cyc & FF & Idt &   FID &   FID$^\ddagger$ \\
    \midrule
    \checkmark & & &              &  61.82 &  24.28 \\
    & \checkmark & &             &  33.73 &  14.12  \\
    & \checkmark & & \checkmark  &  29.94 &  12.81 \\
    & & \checkmark & \checkmark  &   \textbf{25.07} &   \textbf{12.03} \\
    \bottomrule
  \end{tabular*}

\end{table}
}

\newcommand{\fakepar}[1]{\noindent\textbf{#1}}
\def\cG{{\mathcal G}}

\def\cR{{\mathcal R}}
\def\cD{{\mathcal D}}

\usepackage{xspace}
\newcommand*{\eg}{e.g.\@\xspace}

\newcommand*{\etal}{et al.\@\xspace}

\makeatletter
\newcommand*{\etc}{%
    \@ifnextchar{.}%
        {etc}%
        {etc.\@\xspace}%
}
\makeatother

\ifCLASSINFOpdf
\else
\fi
\begin{document}
\title{SelFSR: Self-Conditioned Face Super-Resolution in the Wild via Flow Field Degradation Network}
%
%
%
\author{
Xianfang Zeng$^*$,
Jiangning Zhang$^*$,\thanks{$^*$Equal contribution.}
Liang Liu,
Guangzhong Tian,
and Yong Liu$^{\dag}$ Member, IEEE
\thanks{X. Zeng, J. Zhang, and L. Liu are with Institute of Cyber-System and Control, Zhejiang University, Hangzhou 310027, China (email: zzlongjuanfeng@zju.edu.cn; 186368@zju.edu.cn, leonliuz@zju.edu.cn).}
\thanks{G. Tian is with Ningbo Research Institute, Zhejiang University, Ningbo 315000, China (email: gztian@zju.edu.cn).}
\thanks{Y. Liu is with State Key Laboratory of Industrial Control Technology, Zhejiang University, Hangzhou 310027, China (email: yongliu@iipc.zju.edu.cn).}
}

\maketitle

\begin{abstract}
  In spite of the success on benchmark datasets, most advanced face super-resolution models perform poorly in real scenarios since the remarkable domain gap between the real images and the synthesized training pairs.
  To tackle this problem, we propose a novel domain-adaptive degradation network for face super-resolution in the wild.
  This degradation network predicts a flow field along with an intermediate low resolution image. 
  Then, the degraded counterpart is generated by warping the intermediate image.
  With the preference of capturing motion blur, such a model performs better at preserving identity consistency between the original images and the degraded.
  %
  We further present the self-conditioned block for super-resolution network.
  This block takes the input image as a condition term to effectively utilize facial structure information, eliminating the reliance on explicit priors, \eg facial landmarks or boundary. 
  Our model achieves state-of-the-art performance on both CelebA and real-world face dataset.
  The former demonstrates the powerful generative ability of our proposed architecture while the latter shows great identity consistency and perceptual quality in real-world images.
  \end{abstract}


%

\section{Introduction}\label{sec:introduction}
\IEEEPARstart{F}{ace} super-resolution (FSR), also known as face hallucination, aims at restoring high-resolution (HR) face images given low-resolution (LR) counterparts.
This fundamental technology can improve the perceptual quality of face images in various scenarios like video surveillance or online meeting. 
Some recognition tasks such as identity recognition or facial analysis also benefit from the enhanced details.
The performance of super-resolution (SR) on benchmark datasets has been improved rapidly with numerous proposed approaches~\cite{wang2018esrgan,zhang2018image,liu2020residual,ma2020deep,menon2020pulse}.
However, most advanced SR networks perform poorly in the real scenarios since the noticeable domain gap between the test data and the artificially synthesized training pairs.

To unbind the limitation of applying SR networks in real-world, various attempts have been made recently. 
Those methods can roughly be divided into three categories: collecting real training pairs with optical zoom~\cite{zhang2019zoom,cai2019toward}, blind super-resolution with kernel estimation~\cite{gu2019blind,zhou2019kernel}, as well as constructing pseudo pairs via deep neural networks~\cite{bulat2018learn,maeda2020unpaired,gong2020learning,wei2020unsupervised}.
Despite being able to capture the real degradation, the collection of real pairs is expensive and laborious.
Moreover, SR networks trained on the collected data are still hard to generalize to the uncovered scenarios.
The second scheme tackles super-resolution in the wild by modeling the unknown downsampling kernel at test phase.
However, those approaches usually assume a parameterized degradation between LR images and HR.
Such an assumption limits the ability to model complicated distortions in real-world.
Hence the third scheme attempts to estimate the real degradation using neural networks.
A degradation network or domain adaptation module is usually utilized to handle complex sensor noise and diverse distortions.

\figintroImg
Falling into the framework of constructing pseudo pairs, our real-world face super-resolution method, named SelFSR,
involves a domain adaptation network in LR space followed by a SR module.
The most distinguished parts of SelFSR come from the displacement-based adaptation network and the self-conditioned SR network.

Unlike previous methods~\cite{maeda2020unpaired,gong2020learning} that directly utilize CycleGAN estimating the real degradation, we propose a displacement-based network to generate noisy LR images.
Our motivation is that CycleGAN-based domain adaptation methods hardly maintain the appearance feature of input images.
For example, the skin color of the input face is usually drifted after the adaptation processing.
We argue that this drawback is caused by estimating noisy images directly in the whole RGB space, where the huge output space is easy to introduce color bias into the generated samples.
%
Hence, our adaptation network predicts a flow field along with an intermediate image, instead of a noisy LR directly.
The intermediate image is forced to approximate the input with an identity loss, while the estimated flow field warps the intermediate image to generate the noisy counterpart.
Since the flow field indicates the motion of pixels, the resulting network has a preference on capturing motion noisy in degradation processing and performs successfully at preserving appearance consistency.
As a comparison, The CycleGAN based models have no inductive bias about motion, and usually involves the color drifting into the results. 

For face super-resolution, we propose the novel self-conditioned block to construct our SR network.
In mainstream FSR methods~\cite{ma2020deep,chen2018fsrnet}, explicit priors like facial landmarks are utilized to provide the localization of facial components as well as global structure.
However, this structure information is naturally contained in face images.
For instance, a face image filtered by the Sobel kernel can provide rough facial boundary information~\cite{ma2020structure}.
So we have the following question: \textit{Can we find an effective way to directly utilize the structure information contained in input images?}
To this end, we treat the input image as a condition term containing facial structure information.
The LR image is upsampled to different resolutions and fed to the SR network multiple times.
The proposed self-conditioned block extracts the implicit structure information from the input and modulates network activations.
Qualitative and quantitative (un)supervised results on CelebA dataset demonstrate the powerful generative ability of our architecture, which has no reliance on auxiliary facial priors.

In summary, our contributions are three folds.
(i) We propose a novel way to perform domain adaptation utilizing a displacement-based network. Such a model performs successfully at identity consistency since its preference for capturing the motion noise.
(ii) We provide a novel backbone for FSR task, which outperforms previous state-of-the-art methods, with fewer parameters, and no requirement of explicit facial priors.
(iii) Experimental results on real face dataset indicate that our approach can generate high-quality and identity-consistent FSR results on real-world images.

\section{Related work}
\fakepar{Face generation.}
Recently, some methods can directly generate realistic faces from the latent space, which boosted many applications in face editing~\cite{zeng2020realistic,shen2020closed,zhang2020freenet} and face super-resolution~\cite{menon2020pulse}.
Tero~\etal~\cite{karras2017progressive} described a progressive growing methodology for the face generation on 1024$\times$1024 resolution from an underlying code. 
StyleGAN~\cite{karras2019style} proposed a style-based generator that manipulates the activations of image features at different scales, synthesizing extremely naturalistic face images.
Tero~\etal~\cite{karras2020analyzing} further analyzed the influence of instance normalization and improved the image quality via path length regularization and modified architecture.
As face super-resolution is a case of conditional generation processing, researchers can benefit from the advanced methods in face generation.
Inspired by the remarkable success of the style-based architecture in face generation~\cite{karras2019style,karras2020analyzing,park2019SPADE}, we proposed the self-conditioned block which takes the input as a condition term.
Unlike some methods~\cite{korshunova2017fast, banerjee2020hallucinating, lai2018fast} which concatenate the input images onto intermediate feature maps, our self-conditioned block modulates the activations of each layer given the input.

\fakepar{Image super-resolution.}
The single image super-resolution performance on benchmark datasets has been boosted continuously by numerous proposed models~\cite{wang2018esrgan,zhang2018image,liu2020residual,ma2020deep,lai2018fast}, which usually achieve state-of-the-art performance via advanced network architecture or learning strategy.
Some fundamental and classic network designs in SR models contain recursive learning~\cite{kim2016deeply}, residual learning~\cite{he2016deep}, dense connection~\cite{huang2017densely}, multi-path learning~\cite{lim2017enhanced} and attention mechanism~\cite{zhang2018image,7433993,9293182}.
Kim~\etal~\cite{kim2016deeply} proposed Deeply-recursive Convolutional Network (DRCN) to recursively learning high-level representations with a weight-shared module.
In residual learning, the model only predicts the residual map that transforms the LR image into an HR image.
Since the learning difficulty is significantly reduced with residual learning, it is widely used in SR models~\cite{kim2016accurate,li2018multi,tai2017memnet}.
Further, SRDenseNet~\cite{tong2017image} introduced dense block into SR models, utilizing all the previous feature maps to learn the missing high-frequency details in the LR image. 
In multi-path learning based SR models~\cite{lim2017enhanced,li2018multi,han2018image}, the features are transferred to multiple paths for different scale or reception field. Multiple representations are then combined to gain improved performance.
As for attention mechanism, Zhang~\etal~\cite{zhang2018image} proposed the channel attention network for image SR task and largely improve performance.
Further, they proposed non-local attention block~\cite{zhang2019residual} to extract local and non-local representations between pixels for a more effective feature aggregation.

Another important factor is learning strategy like loss function or upsampling framework. 
SRGAN~\cite{ledig2017photo} achieved photo realistic SR results by introducing perceptual loss~\cite{johnson2016perceptual} and adversarial loss~\cite{goodfellow2014generative}. 
ESRGAN~\cite{wang2018esrgan} utilized a relativistic version of adversarial loss~\cite{jolicoeur2018relativistic} and got the state-of-the-art perception-oriented performance at that time, largely boosting the development of paired image super-resolution.
As for upsampling framework, the classical frameworks are pre-upsampling SR~\cite{dong2015image} and post-upsampling SR~\cite{dong2016accelerating,shi2016real}.
Recently, the progressive-upsampling framework~\cite{Yifan2018Fully,8100101} is proposed to reduce the learning difficulty and support for multi-scale SR task.
A detailed and comprehensive survey on SR components can be found in recent image super-resolution reviews~\cite{bashir2021comprehensive,8723565}.
Our SR network also belongs to the progressive-upsampling framework, where ProSR~\cite{Yifan2018Fully} and LapSRN~\cite{8100101} are the most similar model to our SR network.
The difference is that they predict the residual map with respect to the LR progressively, while our model take the input as a condition to modulate feature maps.

As a domain-specific image super-resolution task, face super-resolution models usually use image priors like facial-structured information to achieve high SR performance.
For instance, FSRNet~\cite{chen2018fsrnet} use facial parsing maps and landmarks as priors to achieve face SR task.
Ma~\etal proposed an iterative collaboration framework between face super-resolution and landmark estimation.
Instead of directly utilizing facial priors, we propose the self-conditioned block to use the implicit structure information contained in the input.

\fakepar{Real-world image super-resolution.}
Amount of attempts have been made to tackle the real image super-resolution task. 
By adjusting the focal length of a digital camera, several works collected the training pairs in real scenarios~\cite{zhang2019zoom,chen2019camera,cai2019toward}.
Those methods are able to capture the real degradation with an expensive and laborious collection processing.
Another category of approaches sought for real-world image SR from an algorithmic perspective. 
Jinjin~\etal~\cite{gu2019blind} proposed an iterative kernel correction at the test phase for blind SR.
Ruofan~\etal~\cite{zhou2019kernel} incorporated the blur-kernel modeling in the training phase for real LR image super-resolution.
However, those methods usually are limited by the ability of modeling complicated distortions in real-world.
Several researchers, therefore, turned to utilize deep neural networks to estimate the real degradation~\cite{lugmayr2019unsupervised,yuan2018unsupervised,maeda2020unpaired,bulat2018learn}.
Unlike most existing methods which directly utilize CycleGAN estimating the real degradation, we propose a displacement-based architecture to perform the domain adaptation task.

\section{Proposed Method}\label{method}
Given a set of unpaired real-world LR and HR images $\{(I_{real,i}^L)_{i=1}^M, (I_{real,j}^H)_{j=1}^N\}$, we aim to learn a SR network that upscales the LR images and simultaneously ensures the HR estimation following the real HR distribution.
To this end, we propose a two-stage approach, SelFSR, to perform face super-resolution in the wild.
Figure~\ref{fig:pipeline} illustrates the training pipeline of our method.
The full framework involves two stages: 
\begin{itemize}
  \item \textbf{Domain adaptive degradation.} We first synthesize clean LR images $I_{syn}^{L}$ by downsampling HR images using a bicubic kernel.
  A generator $\cG(\cdot)$ takes $I_{syn}^{L}$ as input to estimate its noisy counterpart $I_{gen}^{L}$, which is generated by warping the intermediate image $\tilde{I}_{syn}^{L}$.
  Meanwhile, $I_{gen}^{L}$ and real LR images $I_{real}^{L}$ are sent to a domain discriminator $\cD_{real}^{L}$ for decreasing the distance of two distributions.
  \item \textbf{Self-conditioned face super-resolution.} We then construct aligned pseudo pairs $(I_{gen}^{L}, I_{real}^{H})$ using estimated noisy LR images and real HR images.
  A self-conditioned SR network $\cR(\cdot)$ is trained with pseudo pairs to perform real-world face super-resolution.
  Similarly, the discriminator $\cD_{real}^{H}$ is utilized in HR space to enhance the texture details of generated high-resolution images $I_{gen}^{H}$.
\end{itemize}

\subsection{Domain Adaptive Degradation}
Super-resolution models trained by clean paired images fails in handling real-world LR images, whose degradation is unknown and more complicated.
To tackle this problem, we introduce the domain adaptation stage to estimate the noisy LR counterparts of HR images.
Unlike previous methods~\cite{maeda2020unpaired,gong2020learning} that directly utilize CycleGAN to transfer the clean LR images to the noisy ones, we propose a displacement-based architecture to perform this task. 
Such an architecture is more likely to capture motion noise and avoid dramatic shifting on identity features (\eg skin color) of input images.
\figpipeline

As can be seen in Figure~\ref{fig:two_approaches_a}, the domain adaptation network predicts a pixel flow field $\mathcal{T} \in \mathbb{R}^{W \times H \times 2}$ as well as an RGB image $\tilde{I}_{syn}^{L}$.
Each element at $\mathcal{T}_{x,y}$ represents the pixel offset $(\delta{x}, \delta{y})$  for the image branch output $\tilde{I}_{syn}^{L}$. 
Namely, the pixel ${I}_{gen}^{L}(x, y)$ in the final output is sampled at the location $(x-\delta{x}, y-\delta{y})$ in $\tilde{I}_{syn}^{L}$, where a generic sampling kernel (\eg bilinear) is applied~\cite{jaderberg2015spatial}.
We denote the sample location $(x-\delta{x}, y-\delta{y})$ as $(x^s, y^s)$, the sampling process can be formulated as
\begin{equation}
  \begin{aligned}
    I_{gen}^L(x,y) = \sum_{u}^{W} \sum_{v}^{H} \tilde{I}_{syn}^{L}(u,v) k(x^s-u; \Phi_x) k(y^s-v; \Phi_y) \text{,}
  \end{aligned}
  \label{eq:grid_sample}
\end{equation}
where $\Phi_x$ and $\Phi_x$ are the parameters of sampling kernel $k()$.

We utilize the flow field degradation network $\cG$ to estimate the noisy LR images.
A domain discriminator $\cD_{real}^{L}$ is used to distinguish the generated LR images and the real.
During the stage of domain adaptation, these modules are constrained by following terms.

\fakepar{Adversarial loss in LR.} We impose an adversarial constraint~\cite{goodfellow2014generative} for matching the distribution of the estimated noisy LR images and the real. The adversarial loss for $\cG(\cdot)$ is defined as
\begin{equation}
  \begin{aligned}
    \mathcal{L}_{adv}^{L}(\mathcal{G}, \mathcal{D}_{real}^L) 
    &= \mathbb{E} \left[\log \cD_{real}^L (I_{real}^{L} ) \right]  \\
    &+ \mathbb{E} \left[\log (1 - \cD_{real}^L (\cG(I_{syn}^L))) \right] \text{.}
  \end{aligned}
  \label{eq:adv_loss_LR}
\end{equation}


\fakepar{Identity loss.} To keep the identity consistency of estimated LR images and the input, we impose the identity restriction on the image branch output. It reduces the L1 distance between $\tilde{I}_{syn}^{L}$ and ${I}_{syn}^{L}$, denoted as
\begin{equation}
  \begin{aligned}
    \mathcal{L}_{idt}(\tilde{I}_{syn}^{L}, {I}_{syn}^{L})
    = \mathbb{E} [\| \tilde{I}_{syn}^{L} - {I}_{syn}^{L} \|_{1} ] \text{.}
  \end{aligned}
  \label{eq:idt_loss}
\end{equation}

\fakepar{Smoothness loss.} For avoiding quickly-changing and self-crossing in the local deformation , we enforce the smoothness constraint on the predicted flow field $\mathcal{T}$. It minimizes the total variation of flow field horizontally and vertically, denoted as
\begin{equation}
  \begin{aligned}
    \mathcal{L}_{s}(\mathcal{T}) 
    = \mathbb{E}[ \left\| \nabla_x\mathcal{T} \right\|_{1} + \left\| \nabla_y\mathcal{T} \right\|_{1} ] \text{.}
  \end{aligned}
  \label{eq:smoothness_loss}
\end{equation}

The full objective for domain adaptation processing is the weighted sum of above terms:
\begin{equation}
  \begin{aligned}
    \mathcal{L} (\cG, \cD_{real}^L) 
    = & \mathcal{L}_{adv}^L
      + \lambda_{idt} \mathcal{L}_{idt}
      + \lambda_{s} \mathcal{L}_{s} \text{,}
  \end{aligned}
  \label{eq:total_loss}
\end{equation}
where $\lambda_{i}$ is the weight of $i$-th constraint term.
\figgenerator

\subsection{Self-Conditioned Face Super-Resolution}
%
Mainstream SR networks usually use the post-upsampling framework, which involves a heavy trunk component (\eg 345 convolution layers in ESRGAN~\cite{wang2018esrgan}) followed by a lightweight upsampling part.
This unbalanced design performs poorly when extended to high scale factor SR task.
As illustrated in Figure~\ref{fig:generator}, we design a balanced backbone for FSR task, assigning similar numbers of convolution layer for each level.
Since the SR operation is segregated into several small upscaling tasks, the learning difficultly of whole model is reduced.
We further utilize the progressive learning strategy~\cite{karras2017progressive} to train the whole network.
Our self-conditioned SR network starts by $\times2$ scale factor, gradually upscales LR images by a step of $\times2$, and finally achieves a high scale factor.
The model capacity is progressively increased during training process.
The resulting network can upscale LR images with different factors, and easily extend to high scale super-resolution tasks.

To provide facial structure information in each level, LR input is upsampled to corresponding resolution, and then fed to the self-conditioned block multiple times.
The architecture of self-conditioned block is illustrated in Figure~\ref{fig:self_conditioned_block}.
It takes an image as condition term to estimate the standard deviation of deep features. 
The estimated standard deviation $\gamma$ is used for element-wise modulating the normalized activation $\bar{f}$, formulated as
\begin{equation}
  \begin{aligned}
  \mu_{n, c}^{l} &=\frac{1}{H W} \sum_{h=1}^{H} \sum_{w=1}^{W} f_{n, c, h, w}^{l} \\
  \sigma_{n, c}^{l} &=\sqrt{\frac{1}{H W} \sum_{h=1}^{H} \sum_{w=1}^{W}\left(f_{n, c, h, w}^{l}-\mu_{n, c}^{l}\right)^{2}+\epsilon} \text{,}\\
  \bar{f}_{n, c, h, w}^{l} &= ( f_{n, c, h, w}^{l} - \mu_{n, c}^{l}) / {\sigma_{n, c}^{l}} \\
  f^{l+1}_{n,c,h,w} &= \gamma_{n,c,h,w}^l \cdot \bar{f}_{n,c,h,w}^l 
  \end{aligned}
  \label{eq:image_condition}
\end{equation}
where $f_{n, c, h, w}^{l}$ is the activation before normalization and $\mu_{n, c}^{l}$ and $\sigma_{n, c}^{l}$ are the mean and standard deviation of the activation in batch $n$ and channel $c$.
This denormalization processing depends on the condition image and vary with respect to the location, which make the implicit structure information play an important role in modulating network activations.
In this stage, a SR network $\cR(\cdot)$ and a discriminator $\cD_{real}^{H}$ are trained with the pseudo pairs $(I_{gen}^{L}, I_{real}^{H})$ and these modules are constrained by following terms.

\fakepar{Adversarial loss in HR.} An adversarial training constraint is imposed on the sets of $(I_{gen}^{H})$ and $(I_{real}^{H})$ to minimize the distance of two distributions.
We choose the hinge version of adversarial loss for $\cR(\cdot)$, denoted as
\begin{equation}
  \begin{aligned}
    \mathcal{L}_{adv}^{H}(\mathcal{D}_{real}^H) 
    = & - \mathbb{E} \left[\min(0, -1+\mathcal{D}_{real}^H(I_{real}^{H})) \right]  \\
      & - \mathbb{E} \left[\min(0, -1-\cD_{real}^H (\cR(I_{gen}^L))) \right] \\
    \mathcal{L}_{adv}^{H}(\mathcal{R}) 
    = & - \mathbb{E} \left[ \cD_{real}^H (\cR(I_{gen}^L)) \right] \text{.}
  \end{aligned}
  \label{eq:adv_loss_HR}
\end{equation}

\fakepar{Reconstruction loss.} To maintain the consistency of image contents between inputs and generated images. We apply reconstruction loss on the input image $I_{real}^{H}$ and the SR result $I_{gen}^{H}$, that gives
\begin{equation}
  \begin{aligned}
    \mathcal{L}_{rec}(I_{real}^{H}, I_{gen}^{H})
    = \mathbb{E} [\| I_{real}^{H} - I_{gen}^{H} \|_{1} ] \text{.}
  \end{aligned}
  \label{eq:rec_loss}
\end{equation}

\figSelfConditionedBlock
\figcomparisonSR
\fakepar{R1 regularization.} For improving the convergence robustness of adversarial training, we adopt the R1 regularization~\cite{mescheder2018training}. This regularization penalizes the discriminator gradients on the true data distribution for better local stability, denoted as
\begin{equation}
  \begin{aligned}
    \mathcal{L}_{r1}(\mathcal{D}_{real}^H)
    = \frac{r}{2} \mathbb{E}_{p(I_{real}^{H})} [\| \nabla D_{real}^H(I_{real}^{H}) \|^2 ] \text{,}
  \end{aligned}
  \label{eq:r1_loss}
\end{equation}
where $r$ is set to 10 in our experiments.
The full objective of super-resolution stage is
\begin{equation}
  \begin{aligned}
    \mathcal{L} (\cR, \cD_{real}^H) 
    & =  \mathcal{L}_{adv}^H
      + \lambda_{rec} \mathcal{L}_{rec}
      + \lambda_{r1} \mathcal{L}_{r1} \text{,}
  \end{aligned}
  \label{eq:total_loss_SR}
\end{equation}
where $\lambda_{i}$ is the trade-off parameters for $i$-th constraint term.

\section{Experiments}

We first evaluate our model on the synthetic distortion under both paired and unpaired data settings.
We then provide a quantitative and qualitative comparison against state-of-the-art methods with realistic degradation.
Finally, an ablation study and visualization are performed to evaluate the effect of each proposed component in SelFSR.

\fakepar{Implementation details.}
We implement our face super-resolution framework in PyTorch~\cite{van2018automatic}. 
All models are trained by Adam optimizer~\cite{Kingma2014AdamAM} with $\beta_1 = 0.9$, $\beta_2 = 0.99$, and batch size of 48.
We use the trick of equalized learning rate~\cite{karras2017progressive} in our generator and discriminator training. The learning rate, therefore, is set to 0.003 without adjustment during training.
%
%
Training our model on CelebA from scratch takes about 24 hours using 4 Tesla V100.

\tabcomparisonSR
\fakepar{Performance Metrics.}
Under the paired training setting, SR results are assessed with the standard fidelity oriented metrics, structural similarity index (SSIM)~\cite{wang2004image} and Peak Signal to Noise Ratio (PSNR). They are transformed to YCbCr space first and computed on the Y channel only.
Since the SSIM and PSNR metrics are known to not consistent with the human perception of image quality,
we also use Fréchet Inception Distance (FID)~\cite{Heusel2017GANsTB} and Learned Perceptual Image Patch Similarity (LPIPS)~\cite{zhang2018unreasonable} to assess the photo-realism of upscaled images.
When the ground truth of HR images is unavailable, we just report the FID value to measure the distribution distance between the high resolution images and generated results.

\subsection{Experiments on Synthetic Distortions}
\fakepar{Datasets.} 
We conduct experiments on widely used CelebA~\cite{liu2015deep} and Helen~\cite{le2012interactive} datasets, following the experimental procedure described by Cheng~\etal~\cite{ma2020deep}.
%
They crop facial region in each image based on estimated landmark~\cite{baltrusaitis2018openface}, and resize it to $128\times128$ pixels.
Then the HR images are downsampled to $16\times16$ LR counterparts with the bicubic kernel.
For CelebA dataset, we use 168,854 images for training and 1,000 images for testing.
For Helen dataset, there are 2,005 images for training and 50 images for testing.
%
Only random horizontal flip is performed as the regular data augmentation in all experiments.

\fakepar{Hyperparameters.} 
In this subsection, we set the weights in Eq.(\ref{eq:total_loss_SR}) as $\lambda_{rec}=150$ and $\lambda_{r1}=3$. 
For the paired setting, we train our generator for 160k iterations. The model starts by $\times 2$ scale factor, and grows up at 20k and 80k steps. The final scale factor is $\times 8$. 
For the unsupervised setting, we extend the training scheduler to 280k iterations and the generator grows up at 40k and 120k steps.
Without the groundtruth of HR images, Eq.(\ref{eq:rec_loss}) is modified to a cycle consistency formulation:
\begin{equation}
  \begin{aligned}
    \mathcal{L}_{rec}(I_{syn}^{L})
    = \mathbb{E} [\| (\cR(I_{syn}^{L}))_{\downarrow}  - I_{syn}^{L} \|_{1} ] \text{,}
  \end{aligned}
  \label{eq:rec_loss_unpair}
\end{equation}
where $(\cdot)_{\downarrow} $ means bicubic degradation. 

\fakepar{Comparison with state-of-the-art methods.}
We show some SR results of different methods in Figure~\ref{fig:comparisonSR}.
It can be seen that our model is able to generate photo-realistic images and synthesize various facial details like: glasses, pupils, wrinkles, beards and teeth.
Among all baselines, DICGAN is most close to our approach and can output plausible images. However, our model outperforms it at detail generation.
For instance, in the fourth row, DICGAN is failed in generating correct details of pupils  as our model does. 
Another example is that the tooth generated by DICGAN looks blur as shown in the second and fourth row.
Compared with other baselines, SelFSR produces much more realistic textures while other methods yield obvious artifacts and distortions:
PFSR is able to produce realistic SR results but sometimes performs poorly at preserving facial structures; 
the SR results created by FSRGAN are over smooth;
DIC achieves bad performance at human perceptual as a PSNR-oriented method.
The qualitative comparison with state-of-the-art face SR methods, especially DICGAN, demonstrates the powerful generative ability of the generator constructed by self-conditioned block.

Following previous works~\cite{chen2018fsrnet,ma2020deep}, we assess the quality of the SR results with the PSNR and SSIM. Since the above metrics are known to not correlate well with human perception, we also report the image quality metrics FID as well as LPIPS.
As shown in Table~\ref{tab:comparisonSR}, our supervised method performs better than DICGAN and largely outperforms other baselines.
Compared with DICGAN, our end-to-end approach does not need auxiliary face landmark information nor iterative collaboration during the training phase.
Namely, our model is more effective (no landmark) and efficient (no iteration) in the training phase.
It can be seen that our approach can achieve high performance at all metrics.
This indicates that SelFSR is able to preserve pixel-wise fidelity while increasing perceptual quality of generated super-resolution images.

\fakepar{Unsupervised results.}
Furthermore, we evaluate our SR network in an unsupervised manner where we assuming the LR and HR images are unpaired.
The ground truth is only used to assess the image quality of SR results.
As shown in Figure~\ref{fig:comparisonSR}, our model can also generate pleasant SR images under the unpaired setting.
The SR results are comparable to supervised variation as well as DICGAN though the corresponding PSNR and SSIM values are almost worst.
Such an observation is reasonable since SSIM and PSNR are fidelity oriented metrics while the ground truth is absent under this setting.
For FID and LPIPS, the unsupervised result only performs slightly worse than the supervised one.
This result indicates a promising potentiality in misaligned scenarios where the LR and HR images come from the same distribution but have an unavoidable misalignment, \eg data captured by a DSLR camera~\cite{zhang2019zoom}.

\subsection{Experiments on Realistic Distortions}
\fakepar{Real-world face image dataset.}
For real-world face super-resolution, we follow a similar data setting in previous works~\cite{bulat2018learn, maeda2020unpaired}.
The authors created a face dataset of 182,866 HR images from CelebA~\cite{liu2015deep}, AFLW~\cite{koestinger2011annotated}, LS3D-W~\cite{bulat2017far} and VGGFace2 ~\cite{cao2018vggface2}.
LR face images were collected from Widerface~\cite{yang2016wider} for diverse degradation and noise.
In total, there are 53,254 LR images for training and 3,000 images for testing.
The HR and LR training images have different distribution, with $64\times64$ and $16\times16$ pixels respectively.

\tabcomparisonRealSR
\figcomparisonRealSR
\fakepar{Hyperparameters.}
During the domain adaptation stage, we set the hyperparameters in Eq.(\ref{eq:total_loss}) as $\lambda_{idt}=10$ and $\lambda_{s}=1$. 
we train the adaptation network for 180k iterations with a learning rate of 2e-4, which linearly decays to zero over the last half steps.
For the SR network, the weights in Eq.(\ref{eq:total_loss_SR}) is set as $\lambda_{rec}=150$ and $\lambda_{r1}=3$. 
The SR network is trained for 160k iterations, starting with a factor of $\times 2$ and growing up at 60k steps. The final scale factor is $\times 4$.

\fakepar{Comparison with state-of-the-art methods.}
Table~\ref{tab:face_sota} shows the quantitative comparison with the related methods.
Our model achieves a slightly better result than Bulat~\etal's and largely exceeds other baselines.
As can be seen in Figure~\ref{fig:comparisonRealSR}, we further provide some perceptual comparisons to two closest baselines~\cite{bulat2018learn,maeda2020unpaired}.
All three approaches can generate plausible results. 
Compared to the model of Shunta~\etal, our model is more inclined to generate smooth results. For instance, we can see intense local noise in their result at the third column.
For the baseline of Bulat\etal, our model outperforms it at maintaining the features of the input, \eg, we can keep the skin color of the LR images better.

\tabablation
\figablation
\subsection{Ablation Study}
\fakepar{Components in domain adaptation stage.}
We conduct the ablation study for domain adaptation on real face dataset.
Visual samples of different variations are displayed in Figure~\ref{fig:comparisonAblation}.
We first assess the performance of our vanilla variation where the model is trained with clean pairs and directly evaluated on real LR images.
As shown in the second column, there are noticeable distortions about facial structure in SR results since the domain gap can not be ignored.
We further study the effect of proposed flow field architecture (FF) by replacing it with CycleGAN.
Comparing the results of columns three through five, we can find that FF outperforms CycleGAN at maintaining the identity characteristic of the input.
%
The quantitative assessment is demonstrated in Table~\ref{tab:ablation}.
We can see that the performance of super-resolution on real face is dramatically improved with adaptive degradation.
For two domain adaptation networks, FF get better FID value than CycleGAN on both the final SR results and the intermediate translation results.
We believe this improvement is benefited from estimating degradation via flow field.
As shown in Figure~\ref{fig:disturb}, we visualize various noisy LR images via disturbing estimated flow field.
It is worth noting that there are no dramatic changes of identity features in all disturbed images,
indicating a pleasurable property of flow field network at identity consistency. 
%

\figDisturb
\tabablationSR
\fakepar{Components in SR network.}
We further conduct an ablation study to evaluate the importance of each component in SR network under both supervised and unsupervised settings.
Our vanilla variation concatenates the LR input with intermediate features instead of using it to modulate network activation.
As shown in the upper half of the Table~\ref{tab:ablation_SR}, all metrics are gradually increased as the different components are added, meaning that each component is useful for the model.
It is worth noting that self-conditioned block improves SR performance with a large margin, 
indicating that this block is  a more effective way to fuse structure information contained in the inputs.

The bottom half shows the results under the unpaired setting, and we can come to a similar conclusion as supervised training for LPIPS and FID metrics.
However, unsupervised training seems not friendly for fidelity oriented metrics like PSNR and SSIM, as they decrease in a large margin, e.g. 26.59 to 23.43 (-3.16) for PSNR and 0.7696 to 0.6539 (-0.1157) for SSIM.
It is a reasonable result since the pixel-level constraint is not adopted under the unpaired setting.
It can be seen that the \emph{progressive} strategy plays an important role for unsupervised training, improving several metrics largely.
We believe the performance improvement is benefited from the training stability improved by the progressive training, which makes model learn the overall distribution at beginning and then enhance the details.

\subsection{Qualitative Results}
\fakepar{Visualization of learned structural information.}
As shown in Figure~\ref{fig:visual}, we visualized some feature maps of condition branch the self-conditioned block.
Since the normalized activation is element-wise multiplied by the output of condition branch, the output of the condition branch can be seen as the weight of each position.
Namely, The bright the area is the attention area in the corresponding channel.
In Figure~\ref{fig:visual}, we can see that the shallow feature maps focus on some the low-level image features like boundary, while the deep feature maps show some semantic structure information like mouth or face.
This visualization intuitively displays the learned structure information in self-conditioned block, which we believe is why our model achieved SOTA SR performance even with less parameters.

\figVisual
\figfailure
\fakepar{Visualization of failure case.}
We also analyze some failure cases to make our method more objective.
The failure cases can be typically divided into two categories: the first case is not distinguishable even by the human perceptual system; the second is some difficult samples like over-blurry or containing occlusion.
The first two columns in the Figure~\ref{fig:failure} are not distinguishable by the human eye, let alone by machine. 
The third column is too fuzzy, while faces in the fourth column has an extreme angle and the last column have a non-negligible occlusion. 
All the above problems will reduce the performance of the super-resolution network.

\section{Conclusion}
In this paper, we propose a novel two-stage framework to perform face super-resolution in the wild.
To avoid the dramatic drifting of the identity features of the input, we propose a displacement-based network to estimate the real LR counterparts. 
Such a model performs better at appearance consistency with the preference of capturing motion noise.
Besides, a backbone built on a self-conditioned block is presented for face super-resolution.
The novel block has shown a powerful ability at extracting structure information from faces.
As a result, the backbone achieves state-of-the-art performance, with fewer parameters and no need of facial priors.
The experiment results on CelebA and real face images show high perceptual quality FSR images, demonstrating the powerful generative ability of our proposed architecture.

\section{Acknowledgements}\label{sec:acknowledgements}
This work is supported by the National Natural Science Foundation of China under Grant 61836015 and Grant 61771193.
{
\bibliographystyle{IEEEtranN}
\bibliography{references}
}
\end{document}